\documentclass[10pt,twocolumn,letterpaper]{article}

\usepackage{iccv}
\usepackage{times}
\usepackage{epsfig}
\usepackage{graphicx}

\usepackage{epsfig}

\usepackage{amsmath}
\usepackage{amssymb}
\usepackage{booktabs}
\usepackage{makecell}
\usepackage{multirow}
\usepackage{enumitem}

\usepackage[breaklinks=true,bookmarks=false]{hyperref}

\iccvfinalcopy 

\def\iccvPaperID{****} 

\ificcvfinal\fi

\begin{document}
	
	\title{Restoring Images Captured in Arbitrary Hybrid Adverse Weather \\Conditions in One Go}
	
	\author{Yecong Wan\\
		China University of Petroleum (East China)\\
		{\tt\small yecongwan@gmail.com}
		\and
		Mingwen Shao\\
China University of Petroleum (East China)\\
\and
Yuanshuo Cheng\\
China University of Petroleum (East China)\\
\and
Yuexian Liu\\
China University of Petroleum (East China)\\
\and
Zhiyuan Bao\\
China University of Petroleum (East China)\\
	}
	
	\maketitle
	\ificcvfinal\thispagestyle{empty}\fi

	\begin{abstract}
		Adverse conditions typically suffer from stochastic hybrid weather degradations (e.g., rainy and hazy night), while
		existing image restoration algorithms envisage that weather degradations occur independently, thus may fail to handle real-world complicated scenarios. Besides, supervised training is not feasible due to the lack of a comprehensive paired dataset to characterize hybrid conditions.
		To this end, we have advanced the aforementioned limitations with two tactics: framework and data. 
		First, we present a novel unified framework, dubbed RAHC, to Restore Arbitrary Hybrid adverse weather Conditions in one go.
		Specifically, our RAHC leverages a multi-head aggregation architecture to learn multiple degradation representation subspaces and then constrains the network to flexibly handle multiple hybrid adverse weather in a unified paradigm through a discrimination mechanism in the output space.
		Furthermore, we devise a reconstruction vectors aided scheme to provide auxiliary visual content cues for reconstruction, thus
		can comfortably cope with hybrid scenarios with insufficient remaining image constituents. Second, 
		we construct a new dataset, termed HAC, for learning and benchmarking arbitrary Hybrid Adverse Conditions restoration.
		HAC contains 31 scenarios composed of an arbitrary combination of five common weather, with a total of $\sim\!316K$ adverse-weather/clean pairs. 
		Extensive experiments yield superior results and establish new state-of-the-art results on both HAC and conventional datasets.
		
	\end{abstract}
	
	\vspace{-10pt}
	\section{Introduction}
	
	\begin{figure}[t]
		\begin{center}
			\includegraphics[width=\linewidth]{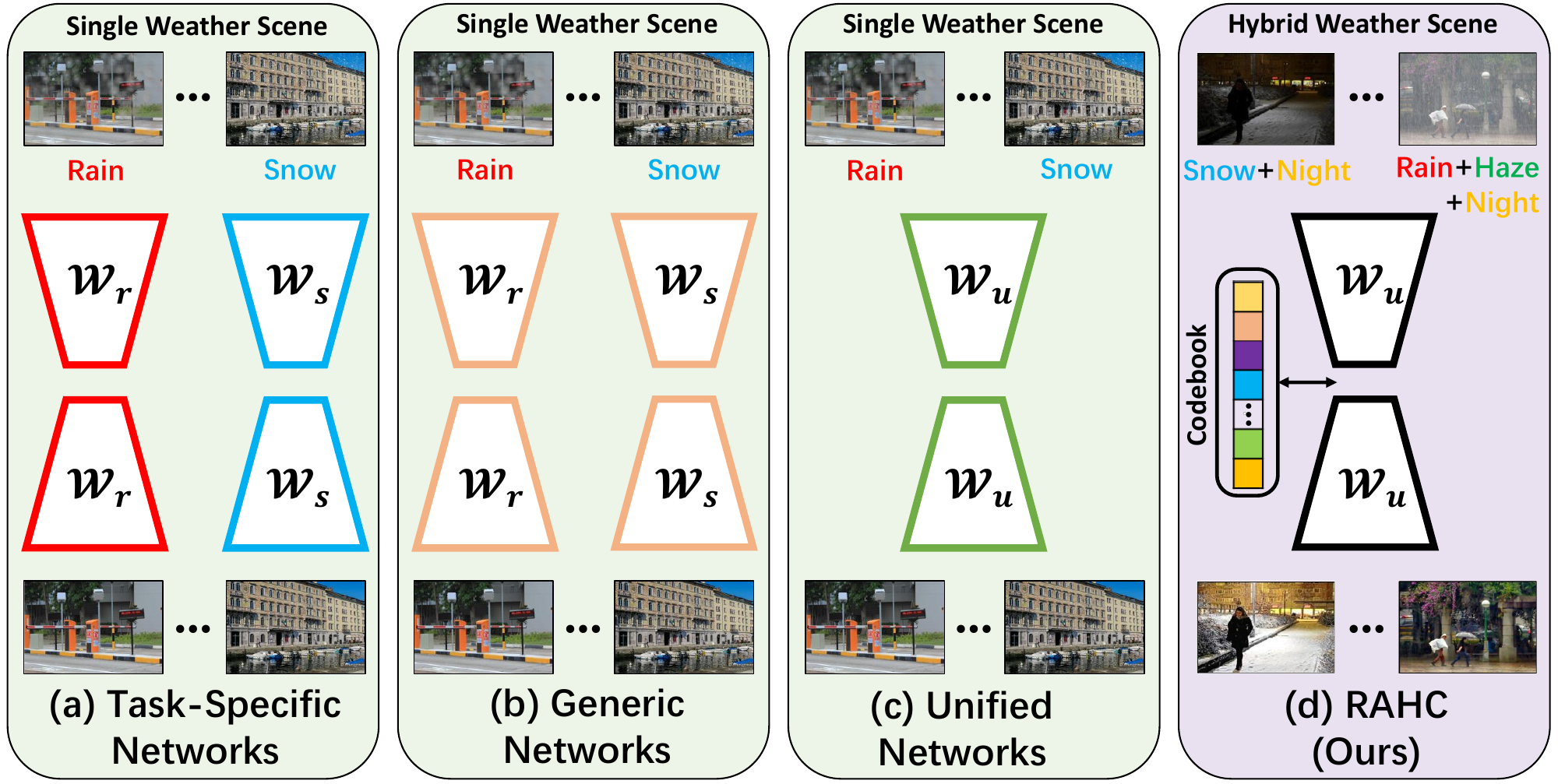}
		\end{center}
		\caption{ \textbf{Overview of adverse conditions restoration frameworks.} (a) Separate networks designed for specific tasks; (b) generic networks with task-specific weights; (c) unified all-in-one networks with single trained weights; (d) our proposed RAHC framework. In contrast to existing approaches that aim to tackle conditions with a single weather type, RAHC can handle arbitrary hybrid adverse weather conditions in one go, thus enjoying better flexibility and practicality in practical applications.}
		\label{figure1}
		\vspace{-10pt}
	\end{figure}
	In real-world adverse scenarios, different weather degradations often occur simultaneously in an uncertain way, e.g., snowy night, rainy night.
	Images captured under these conditions inevitably suffer abysmal visibility and corrupted feature remnants, and the stochastic hybrid degradation may dramatically hamper high-level vision tasks \cite{carion2020end,chen2018encoder,cordts2016cityscapes,long2015fully} in applications such as autonomous driving systems \cite{liang2018deep,prakash2021multi,qi2018frustum} and surveillance systems \cite{perera2018uav}. Unfortunately, prevalent algorithms \cite{zhang2019kindling,liu2019griddehazenet,zhang2018density,quan2019deep} deal with each weather degradation individually ignoring the hybrid degradation characterization of combined action. Specifically, early researchers focused on developing task-specific restoration algorithms for given weather conditions \cite{chen2021all,qian2018attentive,dong2020multi,wang2020model,liu2021retinex} (Fig. \ref{figure1}(a)). Afterwards various generic networks were proposed to handle different degradations with identical architecture \cite{zamir2022restormer,chen2022simple,liang2021swinir,wang2022uformer} (Fig. \ref{figure1}(b)), but with different weights for different tasks. To address this issue, recent researches aim at restoring images under multiple adverse conditions with a single unified model \cite{chen2022learning,li2020all,li2022all,valanarasu2022transweather} (Fig. \ref{figure1}(c)). Nevertheless, these methods still fail to address hybrid adverse weather conditions. Despite few works \cite{li2019heavy,quan2021removing,wan2022image} have partly explored restoring two-weather superimposed scenarios, they were designed for specialized combinations (e.g., rain streak and raindrop) and could not extend to other real-world diverse hybrid scenarios.
	
	To more flexibly and practically deal with real-world scenarios, a turnkey solution that can restore arbitrary hybrid adverse conditions in one go is urgently needed. Compared to this goal, the existing works have the following limitations.  
	(i) Existing restoration networks are limited in characterizing hybrid multiple weather degradations simultaneously due to the lack of a multi-subspace feature extraction mechanism. (ii) Models utilized in single degradation removal are restricted in restoring hybrid adverse weather conditions with insufficient remaining background constituents.
	(iii) Previous unified learning strategies are designed for non-overlapping degradations and are constrained to popularize to diverse hybrid scenarios.

	To tackle the aforementioned problems, in this paper, we propose a novel unified framework RAHC to restore arbitrary hybrid adverse weather conditions in one go (Fig. \ref{figure1}(d)). Specifically, we present three tailored designs to overcome the above limitations. (i) Multi-head blend block (MHBB) for multi-weather degradation representation: the multi-head mechanism overriding the blend operator of convolution and attention can provide multiple ``representation subspaces” \cite{vaswani2017attention} as well as complementary features for hybrid multi-weather learning. (ii) Reconstruction vectors aided restoration (RVA) for hybrid conditions with limited image constituents retention: the discrete representations encapsulated in the Codebook \cite{esser2021taming} pre-trained on large-scale natural images which we refer to as reconstruction vectors, can provide additional visual content cues to auxiliary the reconstruction of realistic and clean output. (iii) Output space discrimination (OSD) for 
	efficient arbitrary hybrid conditions restoration: we design a simple multilabel-classification discriminator from the output space to force the restoration network to learn degradation-independent repair capabilities which can flexibly cope with diverse hybrid scenarios without any complex strategies or modules. Especially noteworthy is that this protocol can be seamlessly integrated into existing universal image restoration algorithms boosting their performance in the all-in-one multi-weather removal setting.

	Apart from the network, we manage to construct a hybrid adverse weather conditions dataset HAC, which covers $\sim\!316K$ pairs of $2^5\!-\!1\!=\!31$ adverse conditions with five common weather types (namely haze, rain streak, snow, night, and raindrop) arranged in combination except for the clean one. 
	To synthesize sufficient and diverse pairwise data efficiently and at low consumption,
	we develop a powerful generator AdverseGAN to learn from adverse conditions so as to approximate the degradation implicitly rather than expensive manual labeling. Thus, the training set can be automatically generated by AdverseGAN with minimal labor cost.
	To guarantee authoritative evaluation, the test set is meticulously handcrafted by our recruited experts. The domain gap between the training and test sets allows for better evaluation of the generalization ability which is critical for real-world applications, especially
	when the real data is infeasible.
	Comprehensive experimental results substantiate 
	the superiority of the RAHC beyond 
	state-of-the-art restoration methods on both HAC and conventional datasets.
	
	In conclusion, the main contributions are summarized as follows:
	\begin{itemize}[noitemsep]
		\item We propose a novel framework, dubbed RAHC,
		to restore diverse hybrid adverse weather conditions while enjoying the properties of being concise, flexible, and powerful.
		\item We present a multi-head blend block to provide multiple representation subspaces for multi-degeneration learning. Meanwhile, a reconstruction vectors aided restoration scheme is devised for severely deteriorated hybrid adverse conditions, as well as an output space discrimination regime for unified learning of diverse hybrid degradations. 	
		\item A new synthetic dataset HAC for arbitrary hybrid adverse weather conditions restoration is constructed. To the best of our knowledge, HAC is the first to encompass such a wide range of scenarios and provides the most comprehensive benchmark for this task.
		\item Extensive experimental results on HAC and conventional datasets demonstrate the effectiveness, superiority, and robustness of our proposed RAHC. 
	\end{itemize}
	\vspace{-3pt}
	\section{Related Work}
	\vspace{-3pt}
	\subsection{Adverse Weather Conditions Restoration}
	\vspace{-3pt}
	Numerous algorithms have been proposed to recover images captured in adverse weather conditions, e.g., rain \cite{wang2020model,fu2017removing}, haze \cite{qin2020ffa,wu2021contrastive,song2022vision}, snow \cite{liu2018desnownet,chen2021all}, etc. While these methods perform well on the specific weather type, significant performance degradation was observed when migrating to other tasks. Aiming at this limitation, a broad spectrum of research \cite{zamir2020learning,zamir2021multi,zamir2022restormer,chen2021hinet,chen2022simple} has explored generic frameworks to accommodate different degradation types with an identical network. Even so, they still require separate training of independent weights for each task, hindering its generalizability. To repair multiple degradations in an all-in-one fashion, several unified models are proposed \cite{chen2022learning,li2020all,li2022all,valanarasu2022transweather}. However, these approaches ignore that real-world adverse conditions often suffer from superimposed multiple degradations \cite{li2019heavy}, e.g., rain streaks and raindrops \cite{quan2021removing}, rain and nighttime \cite{wan2022image}, etc. In this paper, adverse scenarios with an arbitrary hybrid of five weather types are considered for a total of 31 conditions, and RAHC can cope with all conditions in one go driving arbitrary hybrid adverse conditions restoration in a broad sense.
	\vspace{-3pt}
	\subsection{Pseudo Data Generation}
	\vspace{-3pt}
	Data-driven vision tasks rely heavily on high-quality datasets.	Unfortunately, the labeling and acquiring of data tends to be expensive, challenging and time-consuming. To surmount this bottleneck, several recent efforts \cite{zhang2021datasetgan,yang2020surfelgan,wang2021rain,liu2021shadow} delve into pseudo-data generation by leveraging Generative Adversarial Networks (GAN). For instance, Zhang \emph{et al.} \cite{zhang2021datasetgan} proposed a DatasetGAN to synthesize highly
	realistic images with pixel-wise annotations, the model trained on the synthetic dataset can even surpass the ones trained on the real dataset. Yang \emph{et al.} \cite{yang2020surfelgan} released SurfelGAN, which facilitates the training of autonomous driving models by simulating realistic road scenarios. In addition to high-level tasks, this research boom has also been conveyed to low-level tasks. Wang \emph{et al.} \cite{wang2021rain} exceeded SOTA with only $0.17 \%$ original data and synthesized pseudo-data. Yue \emph{et al.} \cite{yue2020dual} introduced a dual adversarial network to simultaneously tackle noise removal and noise generation. Inspired by CycleGAN \cite{zhu2017unpaired}, Wei \emph{et al.} \cite{wei2021deraincyclegan} generated a pseudo-paired dataset Rain200A by cycle translation, and the models trained on it exhibit better robustness and generalization. Inspired by these pioneer works, we pursue the efficient and inexpensive synthesis of paired data for training by utilizing a deliberate GAN.
	\vspace{-3pt}
	\subsection{Perceptual Image Compression}
	\vspace{-3pt}
	Tremendous success in two-stage image generation \cite{esser2021taming,yan2021videogpt,yu2021vector,razavi2019generating,rombach2022high} based on perceptual image compression has been witnessed. These works compress images into discrete latent 
	vectors in the first stage and synthesize high-quality images leveraging encoded vectors in the second stage. In this paper, we focus on perceptual image compression in the first stage. VQVAE \cite{razavi2019generating} first presents an auto-encoder model to implement multi-scale quantization of images.
	Based on VQVAE, VQGAN \cite{esser2021taming} introduces adversarial  and perceptual objectives to obtain higher compression rates while preserving satisfactory perceptual quality. Furthermore, LDM \cite{rombach2022high} explores different compression rates and different kinds of regularizations, deriving a compression model that is more adept at preserving details. Intrigued by the extensive image priors contained in the discrete latent vectors of image compression, we advocate utilizing these auxiliary priors to guide the restoration, since the discrete vector comprises context-rich visual parts.
	\vspace{-3pt} 
	\subsection{Unsupervised Domain Adaptation (UDA)}
	\vspace{-3pt}
	UDA algorithms \cite{hoffman2018cycada,kang2020pixel,murez2018image,tsai2018learning,luo2021category,luo2019taking,zou2018unsupervised} for semantic segmentation aim to constrain the model trained on the source domain to learn domain-invariant features, thereby generalizing to the target domain. One category of these studies \cite{tsai2018learning,luo2021category,luo2019taking,zou2018unsupervised} was devoted to training a discriminator to distinguish whether the output results come from the source or target domain, while the segmentation network learns domain-invariant knowledge to confuse the discriminator. 
	Motivated by these methods, RAHC interestingly treats the uniform restoration of different degradations as a domain adaptive problem, as will be demonstrated, fooling a degradation type classifier is more straightforward and concise than sophisticated units and training strategies
	in degradation-independent feature learning.
	\vspace{-3pt}
	\section{Methodology}
	\vspace{-3pt}
	\subsection{Overview of the Proposed Method}
	\vspace{-3pt}
	\begin{figure*}[t]
		\begin{center}
			\includegraphics[width=\linewidth]{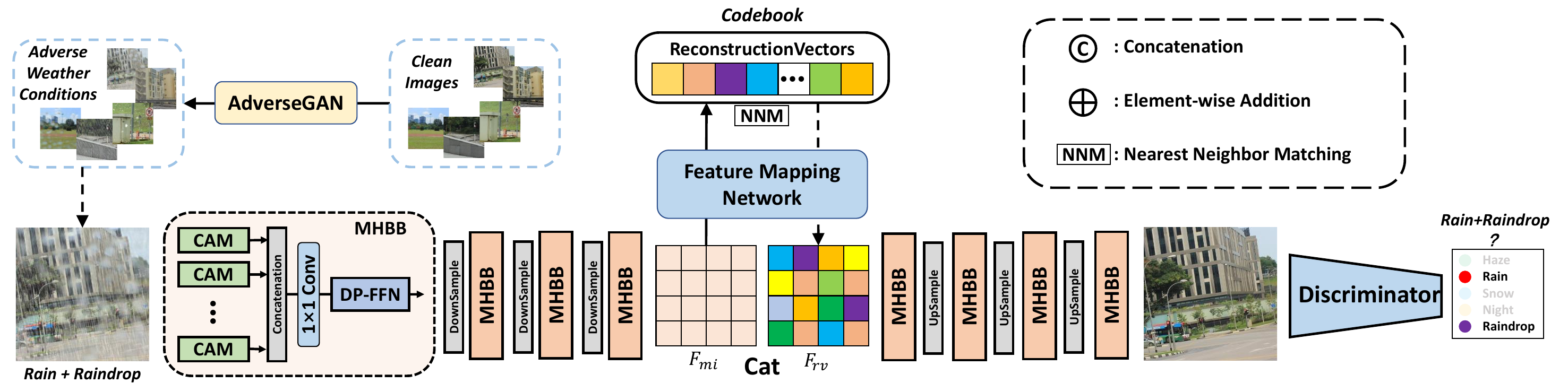}
		\end{center}
		\vspace{-10pt}
		\caption{Illustration of the proposed RAHC architecture. The restoration network consists of an encoder, a decoder, and a feature mapping subnetwork. The mapping network first maps the encoded feature to the latent clean space and then locates the reconstruction vectors in the pre-established Codebook by nearest neighbor matching to provide privileged visual cues for the decoder. To enable degradation-independent learning, we utilize a discriminator to distinguish the type of weather degradation from the restored image while the restoration network struggles to fool the discriminator. }
		\label{figure2}
		\vspace{-10pt}
	\end{figure*}
	\begin{figure}[t]
		\begin{center}
			\includegraphics[width=\linewidth]{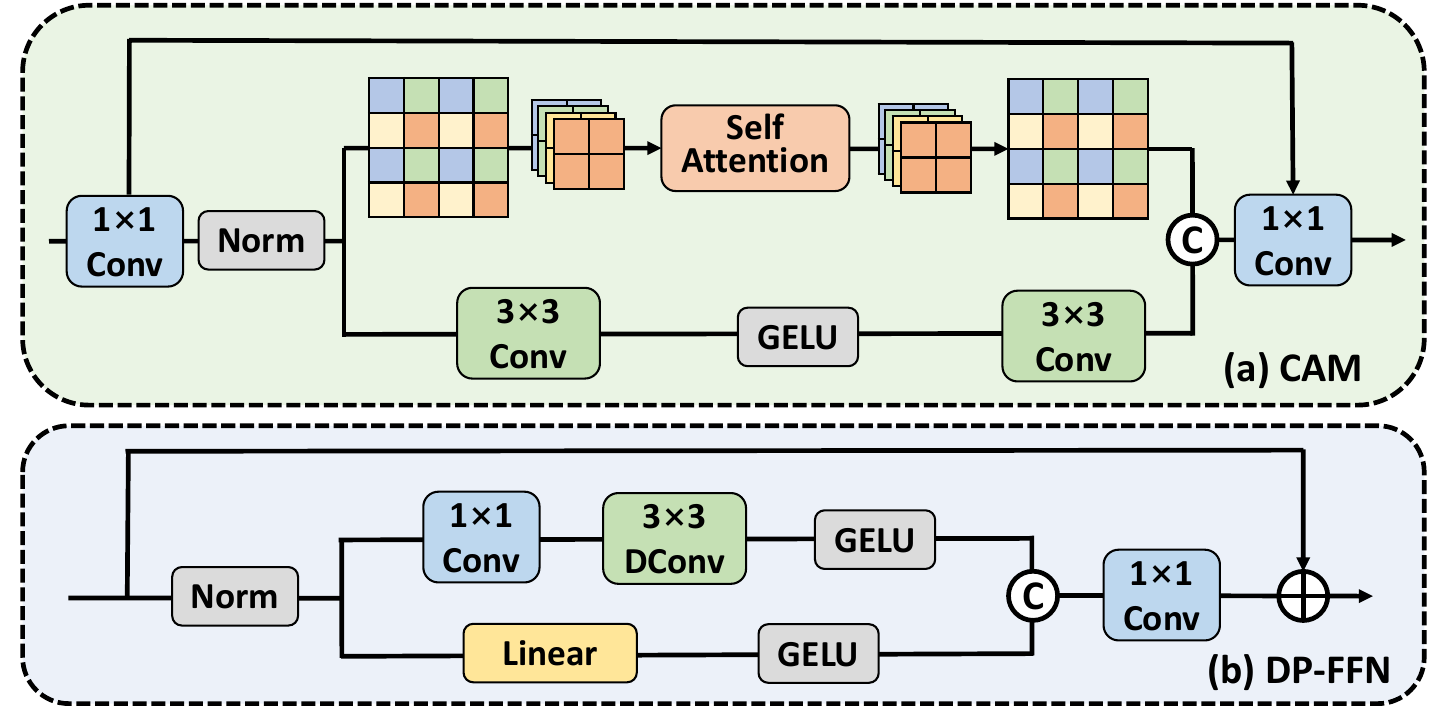}
		\end{center}
		\vspace{-7pt}
		\caption{(a) Schematic diagrams for Convolution-Attention Module (CAM); (b) Dual-Path Feed-Forward Network (DP-FFN). }
		\label{figure2_plus}
		\vspace{-14pt}
	\end{figure}
	The training procedure for RAHC is illustrated in
	Fig. \ref{figure2}. RAHC aims to tackle arbitrary hybrid adverse conditions restoration via a unified framework, and the central notion is to efficiently generate hybrid adverse scenarios by AdverseGAN and then train a degeneration-independent restoration network through multi-head aggregation structure and discriminative learning scheme. Simultaneously, the visual constituents embedded in the Codebook are leveraged to provide auxiliary visual cues for the restoration of highly difficult hybrid conditions. Formally, given a clean image $C \in \mathbb{R}^{H \times W \times 3}$, AdverseGAN first generates the corresponding degraded ones $D \in \mathbb{R}^{H \times W \times 3}$ to obtain the degraded-clean pair. The generated degraded image is then fed into the restoration network to produce the consistent restoration-independent restored image $R \in \mathbb{R}^{H \times W \times 3}$. Meanwhile, the feature mapping network learns the projection from the encoded feature to corresponding clean embedding to locate the reconstruction vectors in the Codebook, supplying additional auxiliary visual atoms for restoration. Finally, the restored result is input into the discriminator to distinguish which type of degradation the restored image suffered before the restoration. The whole procedure is a snap, and there is no extra inference
	cost or modification during testing besides the restoration network.

	\subsection{Network Architecture} \label{3.2}
	The restoration network adheres to a U-shaped structure, which hierarchically cascades multiple tailored multi-head blend blocks, and the knowledge domain is broadened at the bottleneck supported by reconstruction vectors, thereby leading to more optimal repair results. Mathematically,
	given a degraded image $D \in \mathbb{R}^{H \times W \times 3}$,
	a $3 \times 3$ convolution is first adopted to produce shallow
	feature embeddings $F_{in} \in \mathbb{R}^{H \times W \times C}$. Then, $F_{in}$ is propagated through four encoder layers built upon MHBB to obtain the deep feature $F_{mi} \in \mathbb{R}^{\frac{H}{8} \times \frac{W}{8} \times 8C}$. Thereafter, $F_{mi}$ is transmitted by the feature mapping network to locate the reconstruction vectors $F_{rv} \in \mathbb{R}^{\frac{H}{8} \times \frac{W}{8} \times N_z}$ that most likely can reconstruct the hidden clean image from the Codebook, where $N_z$ represents the dimension of reconstruction vector. Eventually, $F_{mi}$ and $F_{rv}$ are concatenated together and fed into the symmetric decoder to recover the final result $R \in \mathbb{R}^{H \times W \times 3}$ via a $3 \times 3$ convolution. 
	Next, we will describe the core components of the proposed
	restoration network. 
	
	\vspace{-1.5pt}
	\noindent\textbf{Multi-Head Blend Block (MHBB).} 
	Existing feature extraction modules lack a multi-degradation representation mechanism to capture the characteristics of hybrid multiple weather, leading to inadequate feature modeling.
	Besides, the complementary properties of convolution, which has strong local computational capabilities, and Transformer, which is excellent at capturing long-range dependencies, make the hybrid structure a better alternative for feature extraction \cite{zhang2022practical,gulati2020conformer}. To this end, we propose a Multi-Head Blend Block (MHBB) to provide multiple ``representation subspaces” as well as complementary features for multi-weather learning. Fig. \ref{figure2_plus} illustrates the two core units (CAM \& DP-FFN) of MHBB. Instead of combining Transformer and Convolution directly in parallel or series, we treat convolution and self-attention as equivalent micro-level operators. And the multi-head mechanism overriding multiple CAMs can provide multi-degradation representation subspaces for unified learning. To be specific, an input feature $X$, is first divided into ``heads" which bears a resemblance to the vanilla Transformer \cite{vaswani2017attention}. The multi-head design allows separate branches to learn different representations and thus adaptively extract different degradation cues to guarantee the ability of diversity restoration. Each head is then transformed by a Convolution-Attention Module (CAM). CAM contains two branches: the attention path and the convolution path, which are split and merged for parallel processing with the addition of $1\times1$ convolution.
	To reduce the high computational effort of vanilla self-attention while preserving the global computational properties, we adopt the pixel-shuffle \cite{shi2016real} operator in the attention path to diminish the number of tokens while avoiding information loss. The convolution path, on the other hand, consists of two convolutional layers and a GELU activation function. Then, the results of different heads are integrated together with a $1\times1$ convolutional stacked subsequently. Locally relevant information is crucial for image restoration, while the original Feed-Forward Network (FFN) is insensitive and inept at this demand. To compensate for the limitations, we invent a Dual-Path FFN (DP-FFN) to extract local contextual information by introducing convolutional branches parallel with a fully connected layer. Overall, the MHBB process can be expressed as:
	\begin{equation}
		\setlength{\abovedisplayskip}{1pt}
		\setlength{\belowdisplayskip}{1pt}
		\begin{aligned} 
			\hat{Y} = & Cat(CAM_1(X_1),CAM_2(X_2),...CAM_k(X_k)),\\
			Y = & DP\mbox{-}FFN(Conv 1\times1(\hat{Y})) .
		\end{aligned}
	\end{equation}
	\noindent\textbf{Reconstruction Vectors Aided Restoration.}
	Existing image restoration algorithms intend to recover the degraded image from the remaining ambiguous content. In this case, the available features are limited, especially in hybrid conditions, and it is extremely challenging to recover high-quality images with insufficient information. Inspired by two-stage image generation models, which build a Codebook with rich contexts in the first stage and then generate images with encoded discrete vectors in the second stage. We utilize the prior context-rich vectors embedded in the Codebook to assist the network in repairing degraded images, which allows the knowledge domain of restoration to be extended from a single image to the entire vectors repository. For the input feature map $F_{mi}$, the mapping network maps the features of the degraded image to the quantization encodings corresponding to the hidden clean image in Codebook, i.e., the reconstruction vectors. Benefiting from the information-rich image components contained in the reconstruction vectors, the restoration network can better reconstruct high-quality images. 
	
	More precisely,
	we utilize the VQGAN \cite{esser2021taming} pre-trained on OpenImages \cite{kuznetsova2020open} with 8192 quantization encodings as a library of reconstruction vectors.  The VQGAN encoder first produces the quantized encoding $F_{rv}^{c} \in \mathbb{R}^{\frac{H}{8} \times \frac{W}{8} \times N_z}$ of the clean image, and for the degraded image feature $F_{mi}$ encoded by the restoration network encoder, the mapping network learns to predict the   
	possible embedding $F_{rv}^{d}$ that consistent with $F_{rv}^{c}$. 
	Attention and convolution layers are cascaded to construct the mapping network which can be optimized by the following cosine similarity loss.
	\begin{equation}
		\setlength{\abovedisplayskip}{1pt}
		\setlength{\belowdisplayskip}{1pt}
		\begin{aligned} 
			\mathcal{L}_{map} =  \sum_{i=0}^{\frac{W}{8}} \sum_{j=0}^{\frac{H}{8}}[ \frac{1-cos(F_{rv}^{d}(i,j) , F_{rv}^{c}(i,j))}{\frac{W}{8}\times \frac{H}{8}}].
		\end{aligned}
	\end{equation}
	
	We can then obtain $F_{rv}$ by leveraging a subsequent nearest neighbor matching $N\!N\!M(\cdot)$ of each
	spatial encoding $F_{rv}^{d}(i,j) \in \mathbb{R}^{N_z}$ from $F_{rv}^{d} \in \mathbb{R}^{\frac{H}{8} \times \frac{W}{8} \times N_z}$ onto its closest reconstruction vector $rv_k$ in the Codebook.
	The schematic illustration of this process is depicted in Fig. \ref{figure2}.
	
	\vspace{-1pt}
	\subsection{Output Space Discrimination} \label{3.3}
	\vspace{-2pt}
	Existing all-in-one restoration approaches rely on distillation \cite{chen2022learning}, degradation guidance \cite{li2022all}, and querying \cite{valanarasu2022transweather} to gain knowledge of different degradations. Although these methods achieve excellent performance for non-overlapping degradation, they are limited in modeling the joint characterization of hybrid multiple degradations. Our intuition is that regardless of the degradation type image suffered, the restored result should be a degeneration-independent high-quality clean image. Thus, we innovatively treat the unified learning of multiple hybrid degradations as a domain adaptive problem and cultivate the degradation-independent adaptive restoration network via a discriminative adversarial learning scheme. In contrast to Li \emph{et al.} \cite{li2020all} that strives to reserve degradation cues to train multiple feature extractors, we dedicate to back-constraining the restoration network to yield consistent degradation-independent ideal images.
	
	\noindent\textbf{Discriminator Training.}
	Given the restoration output $R = Network(D)$, we forward $R$ to discriminator $Dis$ to distinguish the type of
	weather degradation from the restored image. The training of the discriminator can be regarded as a multilabel classification task, and the cross-entropy loss objective can be defined as:
	\begin{equation}
		\setlength{\abovedisplayskip}{1pt}
		\setlength{\belowdisplayskip}{1pt}
		\resizebox{0.9\width}{\height}{$
			\begin{aligned} 
				\mathcal{L}_{d} =\sum_{i=0}^{n-1}[ -t_i\log Dis(R)_i-(1-t_i)\log (1-Dis(R)_i)],
			\end{aligned}$}
	\end{equation}
	where $n$ denotes the number of degenerate types, $t_i = 1$ if $I$ suffer from degradation $i$ else $t_i = 0$.

	\noindent\textbf{Restoration Network Training.}
	First, pixel-level L1 loss is equipped to allow the restoration results $R$ to approximate the ground truth clean image $C$:
	\begin{equation}
		\setlength{\abovedisplayskip}{3pt}
		\setlength{\belowdisplayskip}{3pt}
		\begin{aligned} 
			\mathcal{L}_{net}^{l1} = \parallel R - C \parallel_1.
		\end{aligned}
	\end{equation}

	Second, to allow restoring degeneration-independent results, i.e., results that can not be recognized by the discriminator what degeneration types suffered, a discriminative loss is employed:
	\begin{equation}
		\setlength{\abovedisplayskip}{1pt}
		\setlength{\belowdisplayskip}{1pt}
		\begin{aligned} 
			\mathcal{L}_{net}^{dis} = \sum_{i=0}^{n} -\log (1-Dis(R)_i).
		\end{aligned}
	\end{equation}

	The ultimate goal is to minimize the L1 loss while allowing the results to approximate consistent degradation-independent clean distribution. Meanwhile, perceptual loss $\mathcal{L}_{net}^{per}$ \cite{johnson2016perceptual} is also employed to weaken the interference of noise from pseudo-data on the training. Thus, the final training objective can be expressed as:
	\begin{equation}
		\setlength{\abovedisplayskip}{4pt}
		\setlength{\belowdisplayskip}{4pt}
		\begin{aligned} 
			\mathcal{L}_{net} = \mathcal{L}_{net}^{l1} + \lambda_{dis} \mathcal{L}_{net}^{dis} +  \mathcal{L}_{net}^{per}, 
		\end{aligned}
	\end{equation}
	where $\lambda_{dis}$ is set to 0.1 to balance relative weight of $\mathcal{L}_{net}^{dis}$.
	
	The proposed paradigm can cope with 31 hybrid weather conditions by relying only on a five-class classification discriminator, whereas existing approaches have to treat 31 scenarios separately, resulting in a sophisticated and redundant training process.
	
	\subsection{Hybrid Adverse Weather Conditions Dataset}
\begin{figure}[t]
		\begin{center}
			\includegraphics[width=\linewidth]{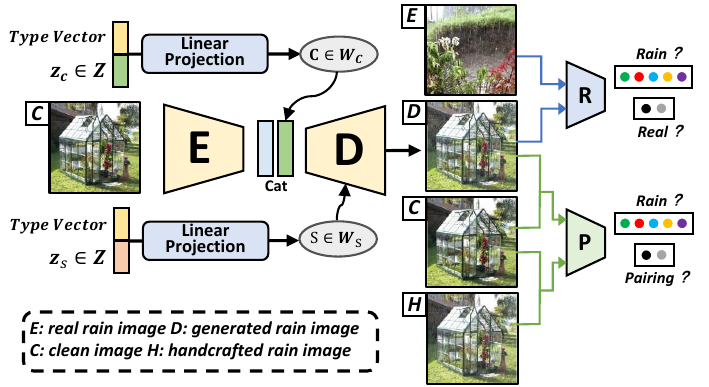}
		\end{center}
		\caption{Pipeline of our proposed AdverseGAN. The generator is an encoder-decoder structure, consisting of two latent space injection branches: content branch (top) and style branch  (bottom). And there are two discriminators: Realism-Discriminator (R) and Pairing-Discriminator (P). }
		\label{figure3}
		\vspace{-16pt}
	\end{figure}

	\noindent\textbf{AdverseGAN for Training Set Generation.} Pseudo-data generation \cite{pan2022gan2x,zhang2021datasetgan,yang2020surfelgan,wang2021rain,liu2021shadow,yue2020dual,wei2021deraincyclegan} has been shown to be feasible and has achieved notable performance. Inspired by the above research, to efficiently generate realistic and controllable paired data for end-to-end training of arbitrary hybrid adverse conditions restoration, we propose to generate pseudo-adverse conditions with GAN rather than artificial processing.
	The key insight is to first train an elaborated AdverseGAN that can generate five basic weather types by leveraging existing data and then mixed-superposition conditions can be generated by recursive calls.
	Therefore, ultimately we can automatically generate 31 adverse conditions without manual operation.
	Schematic illustration of the proposed
	AdverseGAN is provided in Fig. \ref{figure3}.
	
	Following recent works \cite{xu2022transeditor,kwon2021diagonal,alharbi2020disentangled,kim2022style}, our generator is constructed based on a dual space structure, content space $\mathcal{C}$ \cite{alharbi2020disentangled} and style space $\mathcal{S}$ \cite{karras2020analyzing}. The generator first encodes the input clean image into a latent space and then generates the degraded image by injecting randomly drawn content vector $z_c$ and style vector $z_s$ from a normal distribution. The type vector $z_t$, which maps from the type label $t$, is also integrated to control the condition type, similar to the conditional generative adversarial network (cGAN) \cite{van2016conditional}.

	While paired datasets of individual degradation types are readily available, most of them are synthetic and may not adequately simulate real-world degradation scenarios. Additionally, generative adversarial networks may not be able to maintain consistent backgrounds between generated images and their input versions. To address these limitations, we introduce a dual discriminator constraint regime. The realism-discriminator distinguishes between real-world adverse conditions and generated fake images, while the pairing-discriminator distinguishes between real and pseudo pairing data. To enable the generation of multiple adverse conditions, our discriminator produces probability distributions over both sources and condition types \cite{odena2017conditional}. The dual discriminators work collaboratively to ensure that our proposed AdverseGAN generates realistic, content-preserved, and reliable adverse weather conditions.  More details on AdverseGAN architecture and training setting can be found in the Suppl.
	
	\begin{figure}[t]
		\begin{center}
			\includegraphics[width=\linewidth]{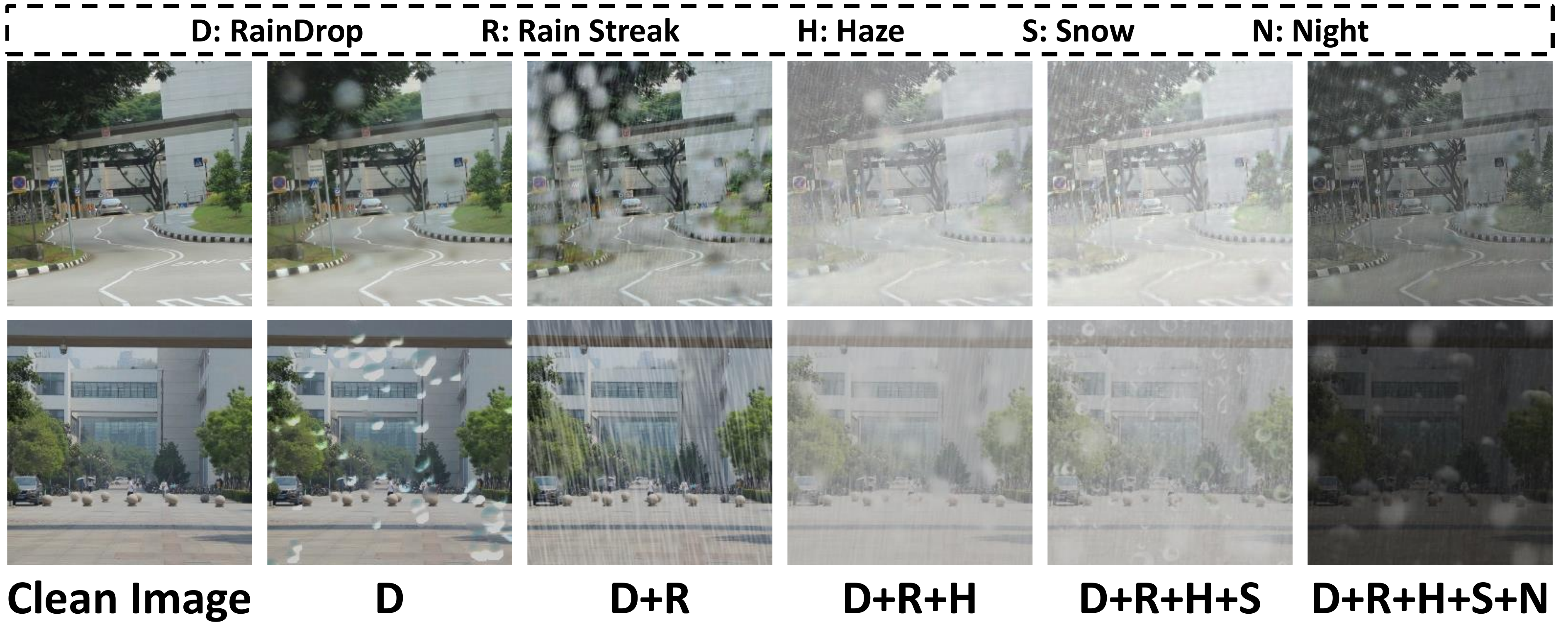}
		\end{center}
		\vspace{-8pt}
		\caption{ Top: Examples of training set images in
			HAC. Bottom: Examples of test set images
			in HAC. Best viewed at screen!}
		\label{figure_dataset}
		\vspace{-10pt}
	\end{figure}
	
	\noindent\textbf{Handcrafted Test Set.} Since the training set can be generated by AdverseGAN, to guarantee the authority and accuracy of evaluation, the test set is carefully fabricated artificially.
	We first capture 200 high-quality well-curated images ($720 \times 480$) as ground truth with Canon EOS 60D, and then synthesize corresponding 31 adverse conditions for each image, resulting in 6200 image pairs.
	The synthesis of haze, rain streak, snow, night, and raindrop follows previous academic literature \cite{li2018benchmarking,fu2017clearing,liu2018desnownet,wei2018deep,quan2021removing}. Photoshop-related works were performed by three photography experts we recruited. Lighting conditions, scenes, subjects, and styles, e.g., are all taken into account to ensure realism and diversity. Examples of HAC are illustrated in Fig. \ref{figure_dataset}.
	
	\vspace{-3pt}
	\section{Experiments}
	\subsection{Evaluation on HAC Dataset}
	\vspace{-3pt}
	\begin{table*}[h]
		
		\caption{Quantitative Comparison with the SOTA methods on our proposed benchmark HAC dataset.
			Top super row: single model instance for single condition (condition-specific). Bottom super
			row: one model instance for all conditions (all-in-one). Multiplicative numeral indicates the number of weather types contained in an image (e.g., Triple represents the average scores of $C_5^3=10$ conditions containing three weather types). Best and second best scores are \textbf{highlighted} and \underline{underlined}.}
		\footnotesize
		\begin{center}
			\begin{tabular}{c|cccccccccc|cc}
				\toprule
				\makecell[l]{\multirow{2}{*}{Method}} &\multicolumn{2}{c}{Single}&\multicolumn{2}{c}{Double}&\multicolumn{2}{c}{Triple}&\multicolumn{2}{c}{Quadruple}&\multicolumn{2}{c}{Pentuple}&\multicolumn{2}{|c}{Average}
				\\
				& PSNR & SSIM& PSNR & SSIM& PSNR & SSIM&PSNR & SSIM&PSNR & SSIM&PSNR & SSIM\\
				\midrule
				\makecell[l]{MIRNet \cite{zamir2020learning}}&28.44  & 0.9285 & 22.10 &0.8749&19.11&0.7940&15.74&0.6483&14.30&0.6257&19.94&0.7743\\
				\makecell[l]{HINet \cite{chen2021hinet}}& 28.99 &0.9393& 23.33 & 0.9101&20.48&0.8422&16.14&0.6982&14.69&0.6689&20.73&0.8117\\
				\makecell[l]{MPRNet \cite{zamir2021multi}}& 28.84 &0.9322  &22.79 &0.9147 &20.19&0.8375&15.99&0.6858&14.65&0.6653&20.49&0.8071\\
				
				\makecell[l]{SwinIR \cite{liang2021swinir}} & 29.08 &0.9411  &24.21 &0.9199 &20.85&0.8517&16.78&0.7599&14.87&0.6672&21.16&0.8280\\
				\makecell[l]{Uformer \cite{wang2022uformer}}& 29.17 & 0.9425 &24.48 & 0.9228&21.05&0.8533&16.89&0.7683&14.95&0.6728&21.31&0.8319\\
				\makecell[l]{Restormer \cite{zamir2022restormer}}& 29.30 & 0.9478 &24.60 & 0.9239&21.08&0.8531&16.92&0.7701&15.04&0.6783&21.39&0.8350\\
				\makecell[l]{NAFNet \cite{chen2022simple}} & \underline{29.35} & \underline{0.9513}&\underline{24.71} &\underline{0.9276}&\underline{21.12}&\underline{0.8583}&\underline{17.01}&\underline{0.7864}&\underline{15.23}&\underline{0.6844}&\underline{21.48}&\underline{0.8416}\\
				\makecell[l]{\textbf{RAHC}} & \textbf{29.45} &\textbf{0.9517} &\textbf{24.98} &\textbf{0.9298}&\textbf{22.03}&\textbf{0.8817}&\textbf{19.53}&\textbf{0.8395}&\textbf{18.09}&\textbf{0.7502}&\textbf{22.82}&\textbf{0.8706}\\
				\midrule
				\makecell[l]{NAFNet \cite{chen2022simple}}& 29.10 &0.9407 &24.22 &0.9143&20.74&0.8465&15.83&0.6623&14.33&0.6287&20.84&0.7985\\
				\makecell[l]{TransWeather \cite{valanarasu2022transweather}}&29.16 &0.9433 &24.25&0.9155&20.83&0.8495&16.72&0.7555&14.72&0.6652&21.14&0.8258\\
				\makecell[l]{AirNet \cite{li2022all}} &29.19 &0.9436 &24.32&0.9207&20.97&0.8522&16.87&0.7655&15.01&0.6741&21.27&0.8312\\
				\makecell[l]{TKL \cite{chen2022learning}}	&\underline{29.23}&\underline{0.9441} &\underline{24.36}&\underline{0.9229}&\underline{21.05}&\underline{0.8531}&\underline{16.93}&\underline{0.7725}&\underline{14.92}&\underline{0.6734}&\underline{21.30}&\underline{0.8332}\\
				
				\makecell[l]{\textbf{RAHC} } &\textbf{29.40}&\textbf{0.9514}&\textbf{24.91}&\textbf{0.9283}&\textbf{22.17}& \textbf{0.8946}&\textbf{19.88}&\textbf{0.8425}&\textbf{18.24}&\textbf{0.7612}&\textbf{22.92}&\textbf{0.8756}\\
				\bottomrule
			\end{tabular} 
		\end{center}
		\label{table1}
		\vspace{-10pt}
	\end{table*}

	\begin{figure*}[h]
		
		\begin{center}
			\includegraphics[width=\linewidth]{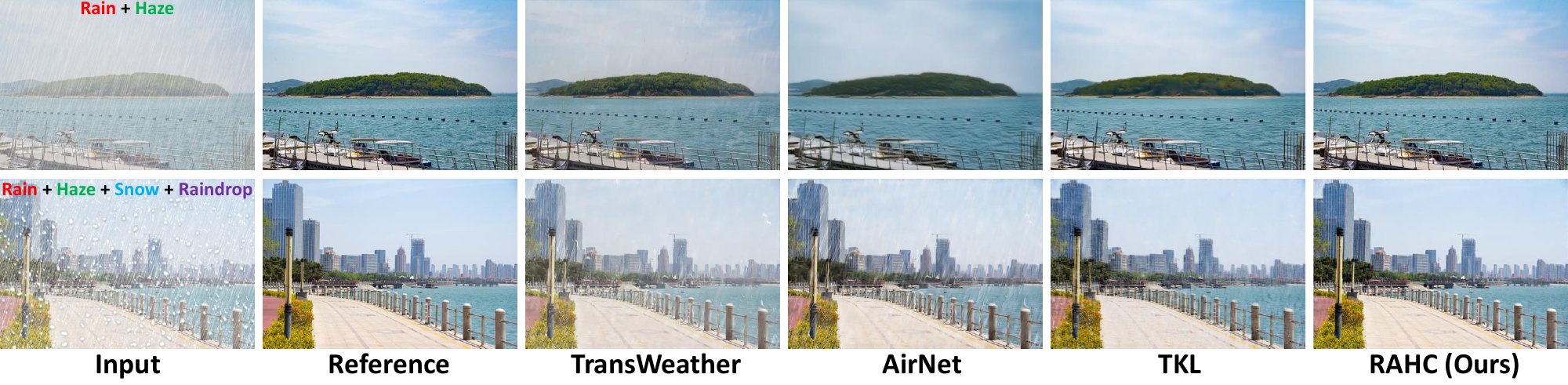}
		\end{center}
		\vspace{-10pt}
		\caption{Visual comparisons with SOTA adverse conditions restoration methods on the HAC dataset. Reconstructed images demonstrate the ability of RAHC to restore arbitrary hybrid adverse conditions while preserving more details, hence yielding more visually pleasing results. \textbf{Best viewed at screen!} }
		\label{figure5}
		\vspace{-12pt}
	\end{figure*}
	
	\begin{table*}[h]
		
		\caption{Quantitative Comparison with SOTA all-in-one multi-weather removal algorithms on five conventional datasets. Due to limited space, more adequate experimental results are given in the Suppl.}
		\vspace{-5pt}
		\footnotesize
		\begin{center}
			\begin{tabular}{c|cccccccccc|cc}
				\toprule
				\makecell[l]{\multirow{2}{*}{Method}} &\multicolumn{2}{c}{Snow100K-L \cite{liu2018desnownet}}&\multicolumn{2}{c}{Raindrop \cite{qian2018attentive}}&\multicolumn{2}{c}{SOTS-outdoor \cite{li2018benchmarking}}&\multicolumn{2}{c}{Rain1200 \cite{zhang2018density}}&\multicolumn{2}{c}{SICE \cite{cai2018learning}}&\multicolumn{2}{|c}{Average}
				\\
				& PSNR & SSIM& PSNR & SSIM& PSNR & SSIM&PSNR & SSIM&PSNR & SSIM&PSNR & SSIM\\
				\midrule
				\makecell[l]{NAFNet \cite{chen2022simple}}&30.32&0.8983&31.94&0.9291&29.87&0.9744&34.12&0.9561&20.16&0.6638&29.28&0.8843\\
				\makecell[l]{TransWeather \cite{valanarasu2022transweather}}&30.28&0.8987&31.74&0.9279&33.01&0.9768&34.05&0.9535&20.42&0.6973&29.90&0.8908\\
				\makecell[l]{AirNet \cite{li2022all}} &30.41&0.9087&32.27&0.9315&33.09&0.9808&34.19&0.9562&20.65&0.6993&30.12&0.8953\\
				\makecell[l]{TKL \cite{chen2022learning}}	&\underline{30.52}&\underline{0.9098} &\underline{32.35}&\underline{0.9341}&\underline{33.13}&\underline{0.9792}&\underline{34.25}&\underline{0.9549}&\underline{21.26}&\underline{0.7131}&\underline{30.30}&\underline{0.8982}\\
				
				\makecell[l]{\textbf{RAHC} } &\textbf{31.08}&\textbf{0.9115}&\textbf{32.82}&\textbf{0.9401}&\textbf{33.54}& \textbf{0.9865}&\textbf{34.71}&\textbf{0.9641}&\textbf{21.56}&\textbf{0.7269}&\textbf{30.74}&\textbf{0.9058}\\
				\bottomrule
			\end{tabular} 
		\end{center}
		\label{table2}
		\vspace{-16pt}
	\end{table*}
	
	The quantitative results on HAC dataset are reported in Tab. \ref{table1}. As can be found, our RAHC delivers unparalleled performance gains and outperforms all competitive models both in the condition-specific setting and in the all-in-one setting, especially
	for extremely adverse scenarios such as ``Quadruple" and ``Pentuple". Notably, RAHC exceeds the top-performing unified approach TKL \cite{chen2022learning} by 3.32dB on PSNR when there are pentuple degradation types. Furthermore, RAHC trained in the all-in-one setting even surpasses the results separately trained in each single condition.
	This phenomenon can be ascribed to the fact that the proposed discriminative learning scheme allows the network to learn more generalized and degradation-independent repair capabilities, and that the reconstruction vectors aided scheme can benefit from more sufficient data. We also demonstrate visual comparisons in
	Fig. \ref{figure5}. As suggested, RAHC recovers clean and crisp results while achieving a harmonious global tone without introducing visible artifacts or color shifts suffered by other methods, especially for complicated hybrid scenarios. 
	\vspace{-3pt}
	\subsection{Evaluation on Conventional Datasets}
	\vspace{-3pt}
	Apart from our proposed HAC dataset, we also conduct experiments with SOTA algorithms on five conventional datasets in the all-in-one multi-weather removal setting. Tab. \ref{table2} shows that our RAHC yields
	consistent and significant performance gains over existing
	approaches on all five weather types. Compared to the recent
	best method TKL \cite{chen2022learning}, our approach achieves 0.44 dB PSNR improvement when averaged across all test sets. 
	\vspace{-3pt}
	\subsection{Ablation Study and Discussion}
	\begin{figure}[t]
		\begin{center}
			\includegraphics[width=\linewidth]{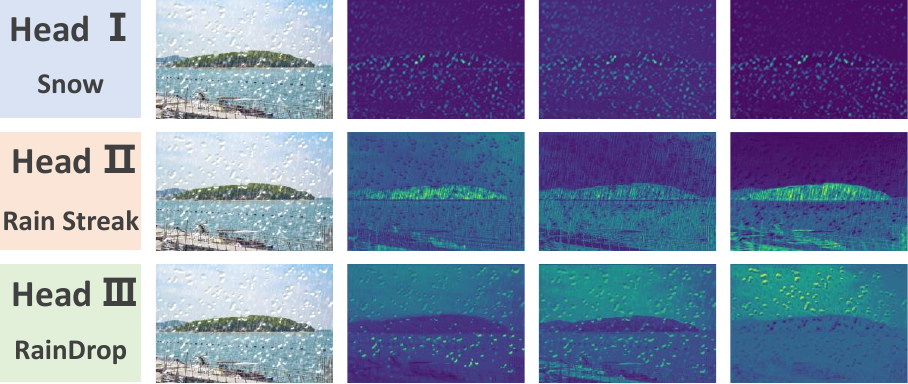}
		\end{center}
		\vspace{-6pt}
		\caption{Visualization of feature maps from different heads in MHBB. The input image contains snow, rain streak and raindrop.}
		\label{figure_mhbb}
		\vspace{-18pt}
	\end{figure}
	\noindent\textbf{Effectiveness of MHBB.} Fig. \ref{figure_mhbb} provides the visualization of feature maps obtained from different heads in MHBB. It is observed that each head in MHBB provides a distinct representation subspace that performs its respective duties to characterize specific degradations under hybrid adverse weather conditions. By leveraging these representation subspaces, MHBB allows more flexibility to characterize the hybrid multiple weather degradations simultaneously.

\begin{figure}[t]
		\begin{center}
			\includegraphics[width=\linewidth]{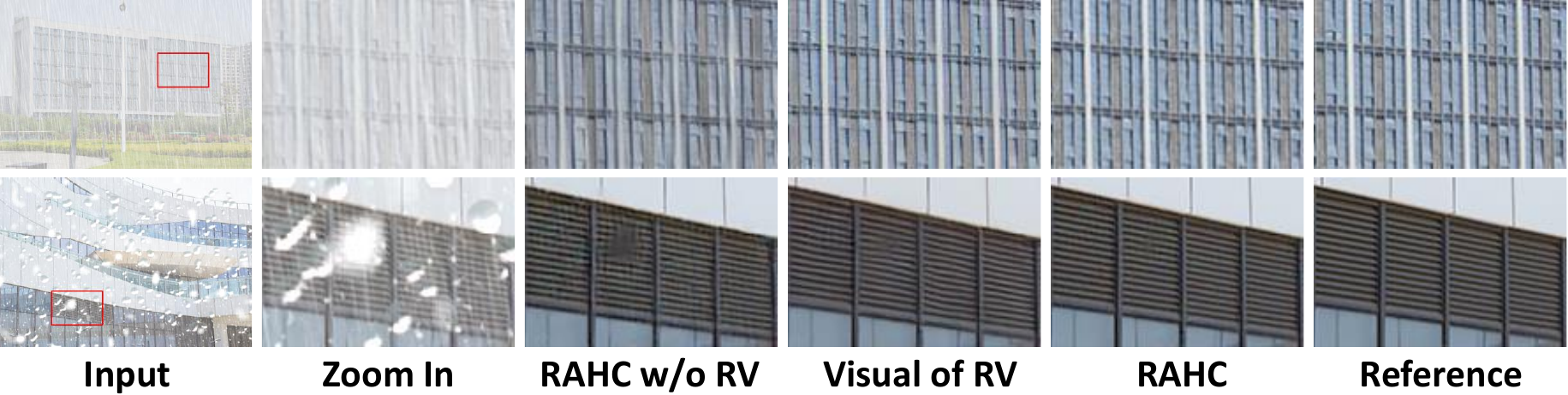}
		\end{center}
		\vspace{-6pt}
		\caption{Visual effect of reconstruction vectors. Visualizations of reconstruction vectors were obtained by the VQGAN decoder. RV represents reconstruction vectors.}
		\label{figure_vqgan}
		\vspace{-18pt}
	\end{figure}

	\noindent\textbf{Visual Evaluation of Reconstruction Vectors.} We also explored the visual effect of reconstruction vectors, and the results are provided in Fig. \ref{figure_vqgan}. As suggested, RAHC without RV tends to produce ambiguous contexts while our RAHC can restore sharper structural and textural details. 
	Naturally, recovering clean images from the reconstruction vectors directly using the VQGAN decoder may be another option, but in this paper, we only leverage the reconstruction vectors as auxiliary features and let the network learn how to utilize them on its own. Such a strategy can also be demonstrated in Fig. \ref{figure_vqgan}. It is observed that the directly restored images (Visual of RV) complement the possible textures of the degraded region but with low fidelity, while our implicit modeling allows the restoration model to use the visual atoms embedded in the reconstruction vector according to its own ``experience", hence restoring more realistic and reliable images with rich details.

	\noindent\textbf{Feature-Level Discrimination vs. Output Space Discrimination.} Analogous to the output space discrimination, we further investigate feature-level discrimination that is input the encoder-extracted features into the discriminator to distinguish the type of degradation. Quantitative experimental results are shown in Tab. \ref{table_fl}, as can be found, both feature-level and output space contribute to learning degradation-independent restoration capacity while our output space is more favored compared to the counterparts. We conjecture that intermediate features containing high-level semantics are more prone to perplex the discriminator, which leads to weaker binding of the discriminator. Especially when performing discriminations at feature-level and output space simultaneously, the higher-level semantics can further disrupt the output space constraints.

	\begin{table}
		\caption{ Quantitative comparison of feature-level (FL) discrimination and output space (OS) discrimination. }
		\footnotesize
		\vspace{-7pt}
		\begin{center}
			\begin{tabular}{ccccc}
				\toprule[1pt]
				Variant& None&FL&FL + OS&	OS (Ours)\\
				
				\midrule[1pt]
				PSNR&22.41&22.71&22.76&22.92\\
				
				\bottomrule[1pt]
			\end{tabular}
		\end{center}
		\label{table_fl}
		\vspace{-15pt}
	\end{table} 
	
	\noindent\textbf{Universal Output Space Discrimination}
	\begin{table}
		\caption{ Results of applying TKL strategy \cite{chen2022learning} proposed by Chen \emph{et al.} (top rows) and output space discriminative learning scheme (OSD) proposed in this study (middle rows) into SOTA universal image restoration methods. Performance gains compared to pure training process are provided in parenthesis. }
		\footnotesize
		\vspace{-4pt}
		\begin{center}
			\begin{tabular}{ccc}
				\toprule
				Method& PSNR &SSIM\\
				\midrule
				Restormer \cite{zamir2022restormer} + TKL &21.14 (\textcolor{red}{$\uparrow$} 0.30)&0.8156 (\textcolor{red}{$\uparrow$} 0.0172)\\
				NAFNet \cite{chen2022simple}  + TKL&21.22 (\textcolor{red}{$\uparrow$} 0.38)&0.8215 (\textcolor{red}{$\uparrow$} 0.0230)\\
				
				\midrule
				Restormer \cite{zamir2022restormer} + OSD&21.31 (\textcolor{red}{$\uparrow$} 0.47)&0.8301 (\textcolor{red}{$\uparrow$} 0.0317)\\
				NAFNet \cite{chen2022simple} + OSD &21.37 (\textcolor{red}{$\uparrow$} 0.53)&0.8339 (\textcolor{red}{$\uparrow$} 0.0354)\\
				\midrule
				TransWeather \cite{valanarasu2022transweather} &21.14&0.8258\\
				AirNet \cite{li2022all}&21.27&0.8312\\
				TKL \cite{chen2022learning}&21.30&0.8332\\
				\textbf{RAHC} &\textbf{22.92}&\textbf{0.8756}\\
				
				\bottomrule
			\end{tabular}
		\end{center}
		\label{table_osd}
		\vspace{-18pt}
	\end{table} 
	As mentioned above, our proposed OSD scheme can be integrated into existing universal image restoration architectures boosting their performance under all-in-one setting. Similarly, TKL \cite{chen2022learning} proposed by Chen \emph{et al.} can also be applied to existing algorithms, and we have compared the superiority and inferiority of the two approaches under a fair backbone. As presented in Tab. \ref{table_osd}, all algorithms show a significant performance improvement under the all-in-one setting when the OSD scheme is equipped which is more obvious and noticeable than the TKL strategy. In particular, Restormer \cite{zamir2022restormer} and NAFNet \cite{chen2022simple} even surpass the extant state-of-the-art unified framework TKL \cite{chen2022learning} with the original backbone in their paper and our mechanism is more flexible and straightforward compared to TKL's multi-teacher distillation. 
	
	\noindent\textbf{Reliability of AdverseGAN.} Inevitably, the generated fake images would introduce undesired artifacts, which seems to compromise the training of the model, but we found the hazard of this to be minimal. To prove this, we randomly take 80\% (a) from the test set of HAC for training and the remaining 20\% (b) for testing. Then based on the clean images in (a), we utilize AdverseGAN to produce the generated data (c) in equal amounts with (a). We train the models on (a) and (c) separately and test them on (b). As shown in Tab. \ref{table_hello}, the models trained on (c) achieve comparable performance to the one trained on (a), while the generated data are labor-effective and can produce infinite unduplicated data.
	\begin{table}
		\caption{ Comparison of training with original data (a) and AdverseGAN generated data. AdverseGAN can generate sufficient and reliable data with extremely cheap costs.}
		\tabcolsep=1.7pt
		\vspace{-4pt}
		\footnotesize
		\begin{center}
			\begin{tabular}{cccc|c}
				\toprule
				Sources&AirNet &TKL &RAHC&Time consuming\\
				\midrule[1pt]
				Original (a)&18.31&18.54&19.16&Approx. 1 month / person\\
				\midrule
				AdverseGAN (c)&18.25 &18.46&19.12&Approx. 10 minutes / GPU \\
				\bottomrule
			\end{tabular}
			
		\end{center}
		\label{table_hello}
		\vspace{-24pt}
	\end{table}

	\subsection{Visual Results on Real-World Conditions}
	\begin{figure}[t]
		\begin{center}
			\includegraphics[width=\linewidth]{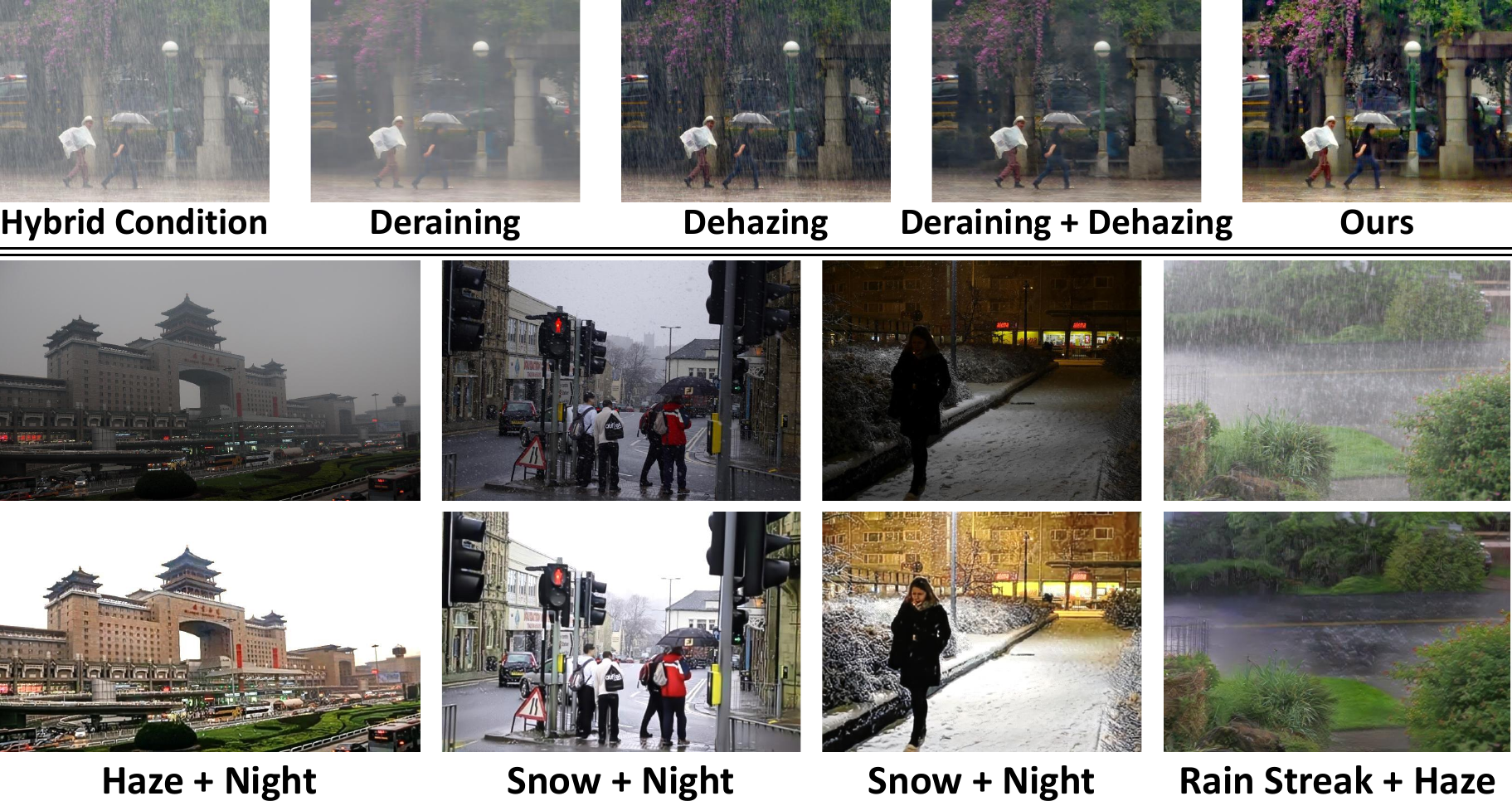}
		\end{center}
		\vspace{-8pt}
		\caption{Five examples of real-world adverse conditions restoration delivered by our RAHC. Best viewed at screen!}
		\label{figure_real}
		\vspace{-13pt}
	\end{figure}
	Fig. \ref{figure_real} exhibits real-world adverse conditions restoration results by RAHC trained on the HAC dataset, the photorealistic results demonstrate both the realistic reliability of
	HAC and the robustness of RAHC. Additionally, this experiment also reveals the truth that real-world scenarios often suffer from multiple superimposed degradations rather than simple one corruption, and our method can restore arbitrary hybrid conditions in one go.
	
	\vspace{-4pt}
	\section{Concluding Remarks}
	\vspace{-3pt}
	In this paper, we proposed a novel unified framework, namely RAHC to restore arbitrary hybrid adverse weather conditions in one go. In contrast to existing frameworks, RAHC can handle severely deteriorated scenarios suffered from hybrid weather degradations and restore arbitrary hybrid conditions with a single trained model by a concise and flexible scheme. Meanwhile, MHBB provides comprehensive degradation characterization and representation support. In addition, we propose a hybrid adverse conditions generation pipeline, based on which sufficient training data can be generated cost-effectively. And then finally established HAC dataset contains $\sim\!316K$ image pairs of 31 types, which richness and diversity render it a competent evaluator.
	Extensive experiments on
	HAC and conventional datasets manifest the effectiveness, superiority, and robustness of our proposed RACH.
	We expect this work to provide insights into arbitrary hybrid adverse conditions restoration and steer future research on this Gordian knot. 
	
	{\small
		\bibliographystyle{ieee_fullname}
		\bibliography{egbib}
	}
\end{document}